\pdfoutput=1
\documentclass[11pt]{article}
\usepackage{acl}
\usepackage{times}
\usepackage{latexsym}
\usepackage[T1]{fontenc}
\usepackage[utf8]{inputenc}
\usepackage{microtype}
\usepackage{multirow}
\usepackage{url}
\usepackage{graphicx}
\usepackage{wrapfig}
\usepackage{amsmath}
\usepackage{amssymb}
\usepackage{color,xcolor,colortbl}
\usepackage{enumitem}
\usepackage{algorithm}
\usepackage{algorithmic}
\usepackage[font={small}]{caption}
\usepackage{bm,bbm}
\usepackage{booktabs}
\usepackage{mathtools}
\usepackage{array}
\usepackage{multirow}
\usepackage{soul}
\usepackage{subcaption}
\usepackage{pbox}
\usepackage{pifont}
\usepackage{arydshln}
\usepackage{boldline}
\usepackage{dsfont}
\usepackage{multirow}
\usepackage{comment}
\usepackage[scaled=0.95]{beramono}
\usepackage{float}
\usepackage{graphicx}
\usepackage{xcolor,colortbl}
\usepackage{todonotes}
\usepackage{hyperref}
\usepackage{longtable}
\usepackage{verbatim}

\definecolor{Gray}{gray}{0.93}
\definecolor{LightCyan}{rgb}{0.9,1,1}
\definecolor{Yellow}{rgb}{1,1,0.9}
\definecolor{Red}{rgb}{1,0.9,1}

\DeclareMathOperator*{\argmax}{arg\,max\,}

\definecolor{darkblue}{rgb}{0.0, 0.0, 0.55}
\newenvironment{fontppl}{\fontfamily{ppl}\selectfont}{\par} 
\setul{0.5ex}{0.3ex} 
\setulcolor{blue} 
\title{
Learning as Conversation:\\
Dialogue Systems Reinforced for Information Acquisition 
}

\author{Pengshan Cai,$^{1}$\thanks{* Work done during internship at IBM Research AI} \ Hui Wan,$^{2}$ \ Fei Liu,$^{3}$ 
\ Mo Yu,$^{2}$ \ Hong Yu,$^{1,4}$ \ Sachindra Joshi$^{2}$ \\
  $^{1}$Manning College of Information \& Computer Sciences, University of Massachusetts, Amherst \\
  $^{2}$IBM Research AI\\
  $^{3}$Department of Computer Science, University of Central Florida  \\
  $^{4}$CHORDS, University of Massachusetts, Lowell \\
  \text{\{pengshancai, hongyu\}@cs.umass.edu}, \ \{hwan, yum\}@us.ibm.com, \\ feiliu@cs.ucf.edu, jsachind@in.ibm.com
  }

\usepackage{amsmath}

\begin{document}
\maketitle

\begin{abstract}

We propose novel AI-empowered chat bots for learning as conversation where a user does not read a passage but gains information and knowledge through conversation with a teacher bot.
Our information-acquisition-oriented dialogue system employs a novel adaptation of reinforced self-play so that the system can be transferred to various domains without in-domain dialogue data, and can carry out conversations both informative and attentive to users. 
Our extensive subjective and objective evaluations on three large public data corpora demonstrate the effectiveness of our system to deliver knowledge-intensive and attentive conversations and help end users substantially gain knowledge  without reading passages. Our code and datasets are publicly available \footnote{Code and data at: \url{https://github.com/IBM/reinforced-dialog-system-for-learning}}
for follow-up research. 

\end{abstract}

\section{Introduction}
\label{sec:intro}
Communication is the central process of education \cite{dewey1923democracy}. In learning as conversation \cite{sharples2005learning}, a student does not read a passage but gains information and knowledge through conversation with a teacher who reads the passage. Compared to the traditional learning by reading, learning as conversation has the advantages that conversation helps students stay engaged and that information is provided piece by piece, which helps strengthen learning with a shorter attention span.

The advantages of learning as conversation have been verified with educational evidence~\citep{mol2008added,lever2011discussing,golinkoff2019language}.
For example, studies have shown that when children read storybooks, parents' guided conversation, e.g., posing questions and providing responsive feedback, substantially amplifies the learning benefits.
While high-quality conversations with experts are not always available, it would be helpful if AI-empowered chat bots could be applied to facilitate users to gain information or knowledge.

\begin{figure}[H]
\centering
\includegraphics[width=3.1in]{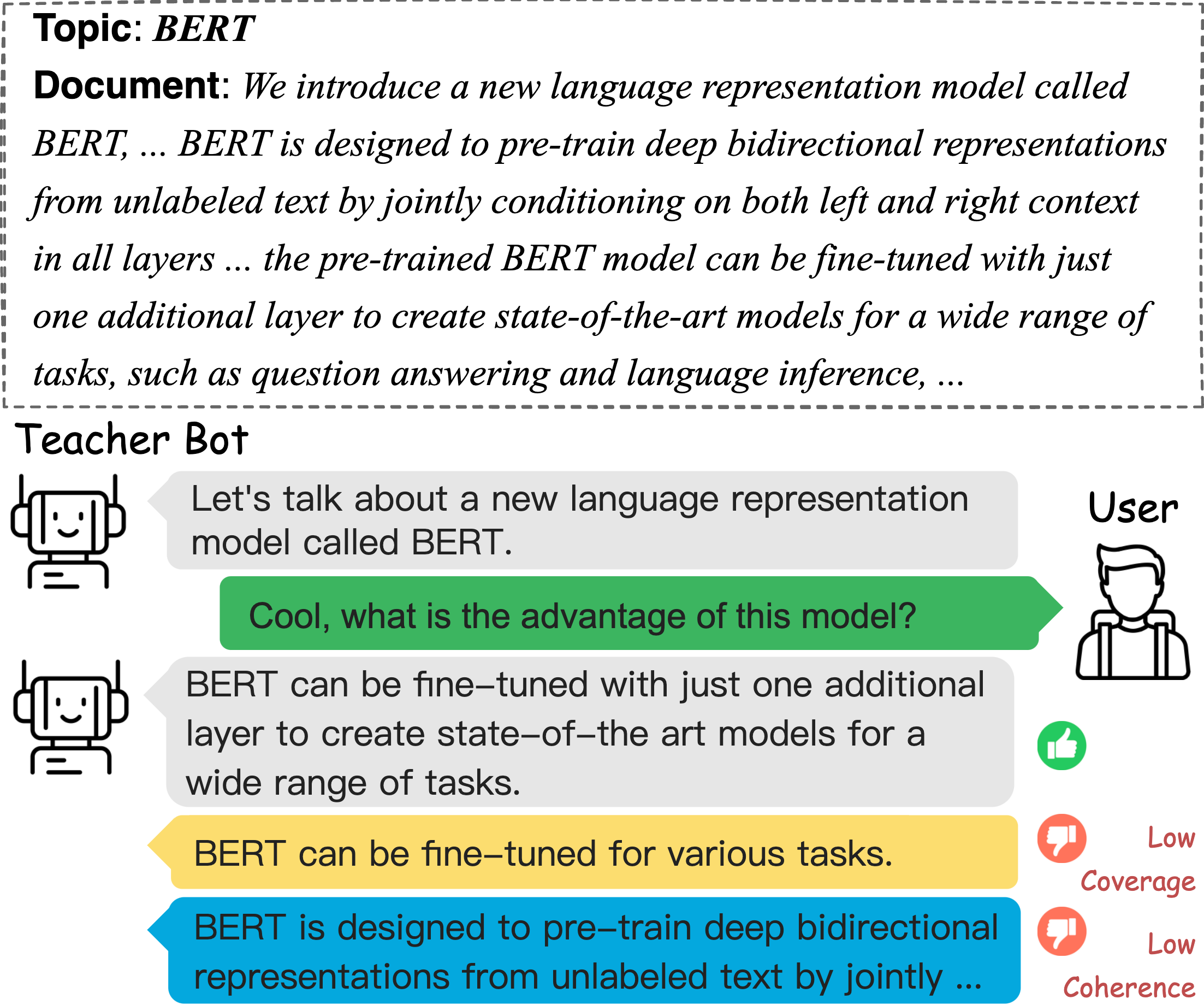}
\caption{The teacher bot educates an user about "BERT" through conversation generated from the document.}
\label{fig:demo}
\end{figure}

In recent years, there has been significant research in content-grounded dialogue generation, where external passages are used to inspire knowledge intensive dialogues. However, these systems or datasets are either for chit chat~\cite{zhou-etal-2018-dataset, dinan2019wizard} or for goal-oriented information seeking~\cite{feng-etal-2020-doc2dial, chen-etal-2021-action}, little work has explored applying chat bots for the learning as conversation purpose.

In this work we propose a novel task for learning as conversation: information-acquisition-oriented dialogue generation. Given a passage, our chat bot actively engages with an end-user  to form a coherent conversation, so that the user could gain knowledge without reading the passage. 
Our task has a broad range of potential application venues in which people traditionally rely on reading to obtain information, including:

\begin{itemize} [topsep=3pt, partopsep=2pt, itemsep=2pt,parsep=2pt]
\setlength\itemsep{-0.3em}
    \item Education: Chat bot helps an user gain knowledge from books or research papers;
    \item News and Media: Through conversation, a user could be provided stories tailored for his/her preference;
    \item Tutorial: While reading an instruction book could be tedious and time-consuming, a chat bot could efficiently walk an user through the process.
\end{itemize}

As shown in Figure~\ref{fig:demo}, for our task, a good conversation should have the following characteristics: 

\begin{enumerate}[topsep=1pt, partopsep=0pt, itemsep=1pt,parsep=1pt]
    \item Coverage: The chat bot should try to convey as much information in the passage as possible, instead of mumbling about irrelevant information;
    \item Coherence: The chat bot's response should be coherent to the user's utterance, making the user feel that his/her questions are followed and addressed. 
\end{enumerate}

In summary, we propose a novel framework which consists of the following two chat bot modules: 1) Teacher bot, which attempts to transfer the information in an input passage to a user through conversation; and 2) Student bot, which responds to Teacher bot to form coherent conversations during training.  
The two bots are trained in a two-phase manner: In Phase 1, we pre-tune the two chat bots on \textsl{Wizard of Wikipedia}~\cite{dinan2019wizard} dataset, enabling both bots with the basic ability of conversing over a passage. In Phase 2, we fine-tune Teacher bot through self-play with Student bot, guided by reinforcement rewards. In this process, we enhance Teacher bot to be more informative while maintaining the ability to coherently address human users. 
Specifically, the fine-tuning phase is unsupervised, i.e. Teacher bot could be transferred to various domains or corpora without additional annotated dialogue datasets. 

Our contributions include: 
1) A novel task of information-acquisition-oriented dialogue system; 
2) A novel unsupervised learning framework which enables a teacher bot to carry out informative and coherent conversations with human users for information acquisition purpose; 
3) Extensive experiments with human evaluation demonstrate the effectiveness of our proposed approach.

\section{Approach}

\begin{figure}[H]
\centering
\includegraphics[width=3.0in]{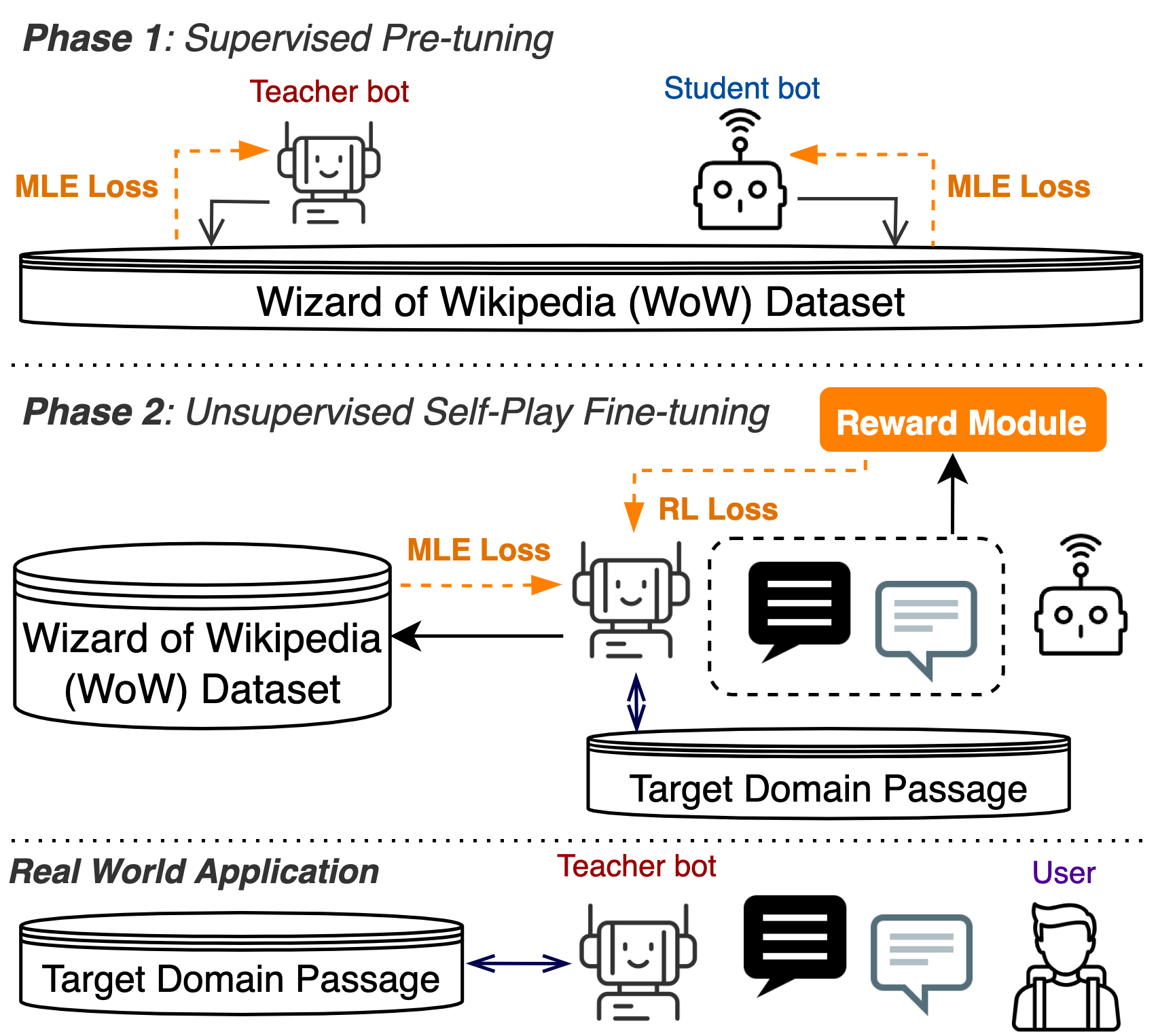}
\caption{Our two-phase training framework}
\label{fig:frame}
\end{figure}

In order to obtain an informative and attentive teaching dialogue system, we propose a framework that consists of two chat bots in different roles, and leverage both supervised learning and unsupervised reinforcement learning, as illustrated in Figure~\ref{fig:frame}. The unsupervised reinforcement learning enables the system to be fine-tuned on other text corpus where no annotation or dialogue data is required.

\subsection{Model Architecture}

Given a passage $P$, the conversation between Teacher bot $X$ and Student bot $Y$ can be denoted as a sequence 
of turns $C = \{U_1^X, U_1^Y, ..., U_{N}^X, U_{N}^Y\}$, where $N$ is the number of turns in the conversation. 
 In order to mimic our use case, Teacher bot has access to $P$ whereas Student bot does not.

\textbf{Teacher bot $X$} aims at transmitting the information in $P$ to the student. 
At the $n_{th}$ turn, $X$ takes as input $P$ and
the conversation history $H_n^Y =\{U_1^X, U_1^Y, ..., U_{n-1}^X, U_{n-1}^Y\}$, and outputs $U_{n}^X$.
Teacher bot $X$ adopts DoHA~\cite{prabhumoye-etal-2021-focused}, a pre-trained model for document-grounded text generation, and is tuned in supervised phase and unsupervised self-play phase.

In order to fine-tune Teacher bot $X$ with reinforcement learning on full conversations, 
as a practical approach, we train a
\textbf{Student bot $Y$} to carry on conversations with $X$. 
Student bot $Y$ takes the conversation history $H_n^Y =\{U_1^X, U_1^Y, ..., U_{n}^X\}$ as input, 
and output $U_{n}^Y$.
It adopts BART~\cite{lewis-etal-2020-bart} model.

\subsection{Phase 1: Supervised Pre-Tuning}

This phase trains Teacher bot $X$ to initialize and carry out conversations based on a given passage $P$, and trains Student bot $Y$ to respond appropriately to $X$. 
To this end, we pre-tune both $X$ and $Y$ on the  \textsl{Wizard of Wikipedia} (\textsl{WoW}) dataset. \textsl{WoW} was chosen as the pre-tuning dataset because of its two characteristics: 1) Open-domain: \textsl{WoW} contains conversations on a broad range of topics and domains across Wikipedia, thus the pre-tuned Teacher and Student bots have greater potentials to be successfully transferred to other domains during the fine-tuning stage; 2) Content-grounded: in \textsl{WoW}, the teaching bot's utterances are grounded on passages, which is similar to our task. 
We present the gold passage to Teacher bot directly, though, different from the \textsl{WoW}'s original setting~\cite{dinan2019wizard} where Teacher bot searches a large corpus for supporting passages.

We optimize the maximum-likelihood objective for Teacher bot by minimizing the following loss:
{\medmuskip=1mu
\thinmuskip=1mu
\thickmuskip=1mu
\nulldelimiterspace=0pt
\scriptspace=0pt
\begin{align*}
L_{mle}^X = -\sum_{n=1}^N\sum_{m=1}^{M_n}log(p(x_{m}|x_{1}, ..., x_{m-1}, H_n^X, P))
\end{align*}}

\noindent where $N$ is the total number of turns in the conversation, $\{x_1, ..., x_{M_n}\}$ is Teacher bot's response at the $n_{th}$ turn, ${M_n}$ is the number of words in $U^X_n$.
The loss function for Student bot, $L_{mle}^Y$, is similar to $L_{mle}^X$, with the exception of not including a passage $P$ as input.
\begin{align*}
L_{mle}^Y = -\sum_{n=1}^N\sum_{m=1}^{M_n}log(p(y_{m}|y_{1}, ..., y_{m-1},  H_{n}^Y))
\label{equ: lms}
\end{align*}

\noindent where $\{y_1, ..., y_{M_n}\}$ is Student bot's response at the $n_{th}$ turn.

\subsection{Phase 2: Unsupervised Self-Play Fine-tuning}
\label{Sec: usp}

In this phase, we aim at improving Teacher bot's ability to present informative and coherent conversations. 
This is achieved by reinforcement learning on Teacher bot with the help of Student bot, and could be applied in a novel target domain even where dialogue dataset is absent.
We adopt a self-play approach, i.e. we let Teacher bot and Student bot chat with each other over a passage in the target domain to generate multiple turns of conversations. In this fine-tuning phase, we keep Student bot frozen, and reward Teacher bot when the generated conversation achieves higher scores.
In order to reduce the variance of the gradient estimate, we apply self-critic reinforcement learning~\cite{Rennie2017cvpr}.
Specifically, at each turn, we let Teacher bot generate two separate utterances: 1) $U^s$, which is sampled from the model, i.e. $x_{m}^s\sim p(x|x_{1}^s, ..., x_{m-1}^s)$, and 2) $U^*$, which is obtained by greedy search, i.e. $x_m^* = \argmax_w p(x|x_{1}^*, ..., x_{m-1}^*)$.
We optimize the model by minimizing the following RL loss:
\begin{align*}
L_{rl}^X = -\sum_{n=1}^N( R(U^s_n) - R(U^*_n) )\sum_{m=1}^{M_n} log(p(x^s_{m}\\|x^s_{1}, ..., x^s_{m-1},  H_{n}^X, P))
\end{align*}

\noindent where $R()$ is the reward function, which we will cover in Section~\ref{Sec:rl}, and $P$ is a passage from the target domain corpus. 

If not taking into account language modeling, optimizing RL loss alone would lead Teacher bot to generate inarticulate and even grammatically incorrect utterances. 
To keep the fluency of Teacher bots, we optimize a combined loss $L^X$ consisting of RL loss $L_{rl}^X$ on the new target domain data and MLE loss $L_{mle}^X$ on the pre-tuning dialogue dataset, so the language style acquired during the pre-tuning phase would not get lost during RL fine-tuning:
\begin{align*}
L^X = \gamma L_{rl}^X + (1-\gamma)L_{mle}^X
\end{align*}

\noindent where $\gamma\in(0,1)$ is a scaling factor accounting for the emphasis on $L_{rl}^X$ and $L_{mle}^X$. 
We note that 
while $L_{mle}^X$ should be obtained on an annotated content-grounded dialogue dataset (e.g. \textsl{WoW}), $L_{rl}^X$ could be obtained on any target domain passage corpus even without dialogue data. This enables our approach to be transferred to an unsupervised text corpus.

\subsection{Reward Functions}
\label{Sec:rl}

\subsubsection{Coverage}

We define the coverage reward of a Teacher bot's utterance $U^X$ as: 
\begin{align*}
R_{cov} = \textsc{Rouge}_1(P, H + U^X) - \textsc{Rouge}_1(P, H)
\end{align*}

\noindent where $\text{Rouge}_1(P, H)$ is the Rouge-1 F1~\cite{lin-2004-rouge} score of the conversation history $H$ to the input passage $P$. Intuitively, this function favors utterances that cover more information in the passage and have less overlap with the conversation history. 

\subsubsection{Coherence}
\label{sec: coh}

\begin{description}[style=unboxed,leftmargin=0cm]
\item[Dialogue coherence datasets]
We explore neural coherence scoring models trained on two open-domain dialogue coherence datasets: 

1. \textsl{WoW-coherence dataset} \quad
We reuse the \textsl{WoW} dataset to heuristically build a dialogue coherence classification dataset. Specifically, for each multi-turn dialogue in \textsl{WoW}, we label the ground truth response to its conversation history as \emph{coherent} response, and all later responses in the same dialogue as \emph{incoherent} responses. 

2. \textsl{InferConv dataset}~\cite{dziri-etal-2019-evaluating} \quad
This is an open-domain dialogue coherence classification dataset built from PersonaChat conversational data~\cite{zhang-etal-2018-personalizing}.
The dataset casts a response as the hypothesis and the conversation history as the premise, thus convert dialogue coherence evaluation into an NLI task. 
The dataset classifies the relationship between the response and the conversation history into three categories: \emph{entailed}, \emph{neutral} and \emph{contradict}.
Table~\ref{Tab:sta2} summarizes statistics of these datasets. 

\item[Coherence scoring models]
Based on the same pre-trained model BERT~\cite{devlin-etal-2019-bert}, we train two different coherence scoring models on the two dialogue coherence classification datasets respectively. 
Both models take the concatenation (with [SEP]) of the conversation history and a candidate response as input, 
and minimize the cross entropy loss between the predicted label and the gold label. 
We use different methods to attain the coherence reward $R_{coh}$ from the two models.

For model \textsl{WoW-coherence}, we define the coherence reward with softmax: 
\[R_{coh} = \frac{e^{o_c}}{e^{o_c} + e^{o_i}}\]
 where $o_c$ and $o_i$ are the logits for \emph{coherent} and \emph{incoherent} labels in the output layer. 
For model \textsl{InferConv}, 
we observe some responses labeled as \emph{neutral} are appropriate responses but are not closely related to conversation history (e.g. \emph{``That's interesting!''}),
 we thus heuristically assign constant scores $s_e$, $s_n$ and $s_c$ as coherence reward $R_{coh}$ when the response is predicted as \emph{entailed}, \emph{neutral} and \emph{contradict}.
In the remainder of the paper, we use \textsl{WoW-coherence} as the default coherence model, and compare it with \textsl{InferConv} in Section~\ref{sec:obj_results}.

\end{description}

\subsubsection{Mixed Reward}

The coverage and coherence rewards are combined with a hyper-parameter $\beta$, yielding the final reward:
\begin{align*}
    R = \beta R_{cov} + (1-\beta)R_{coh} 
\end{align*}

\section{Experimental Settings}
\label{sec:settings}

We proceed by describing our datasets, comparison systems and evaluation metrics.
We then show the performance of our proposed approach compared to state-of-the-art in \S\ref{sec:obj_results}.

\subsection{Datasets}
\label{sec:datasets}

\textsl{Wizard of Wikipedia}~\cite{dinan2019wizard} contains a total of 22,311 human-human conversations crowdsourced via Amazon's Mechanical Turk. 
The conversations are grounded in Wikipedia passages covering a wide range of topics: \emph{e-book}, \emph{toga party}, \emph{armadillo}, etc. 
Both Teacher and Student bots are pre-tuned on the \textsl{WoW} dataset during Phase 1.
Different from \textsl{WoW}'s original setting,
we present the gold passage to the Teacher bot directly,
instead of searching a large corpus for supporting passages. 
This allows us to focus less on retrieval and more on creating a Teacher bot to deliver informative and attentive dialogues.

We consider knowledge sources of various sorts as Teacher bot's target domain during fine-tuning. 
CNN/DailyMail contains a large collection of online news articles with an average of 781 tokens per article~\cite{see-etal-2017-get}. 
The full content of the article cannot be conveyed in a short conversation. 
Thus, we use the first 130 tokens of each article as a supporting passage, assuming it covers the most important content of the news article.

Academic papers have become an omnipresent source of knowledge.
We create our own dataset containing papers published in recent years (2017--2021) 
at major venues, including ACL, EMNLP, NAACL, EACL, Findings and ICLR conferences.
Similarly, we use paper abstracts as supporting passages instead of full articles. 
Moreover, we include Wikipedia passages from the \textsl{WoW} dataset, without conversations, as another source of knowledge.
The \textsl{CNN-DM}, \textsl{Paper Abstracts} and \textsl{Wikipedia} datasets are used in Phase 2 of unsupervised self-play fine-tuning.  
Statistics of these datasets are summarized in Table~\ref{Tab:sta1}.

\begin{table}[t]
\setlength{\tabcolsep}{4pt}
\renewcommand{\arraystretch}{1.05}
\footnotesize
\centering
\begin{tabular}{|p{2.3cm}rrr|}
\hline
\rowcolor{gray!10}
\textbf{Pre-Tuning} & \multicolumn{2}{r}{\textsl{WoW-Train}} & \textsl{WoW-Valid} \\
\#Utterances & \multicolumn{2}{r}{166,787} & 17,715 \\
\#Dialogues & \multicolumn{2}{r}{18,430} & 1,948  \\
\#Words/utterance & \multicolumn{2}{r}{16.6} & 16.6  \\
\#Words/passage & \multicolumn{2}{r}{110.3} & 109.8 \\
\hline
\hline
\rowcolor{gray!10}
\textbf{Fine-Tuning} & \textsl{Wikipedia} & \textsl{CNN-DM} & \textsl{Paper Abs.}  \\
\#Passages        & 50,000   & 50,000   & 22,512  \\
\#Words/passage & 111.7    & 129.8    & 149.3  \\
\hline
\hline
\rowcolor{gray!10}
\textbf{Test Set} & \textsl{Wikipedia} & \textsl{CNN-DM} & \textsl{Paper Abs.}  \\
\#Passages            &  1,000    &  1,000    &  500      \\
\#Words/passage     &  112.7   &  129.9   &  148.1   \\
\hline
\end{tabular}
\caption{Datasets statistics.}
\label{Tab:sta1}
\end{table}

\begin{table}[t]
\setlength{\tabcolsep}{4.2pt}
\renewcommand{\arraystretch}{1.05}
\footnotesize
\begin{tabular}{|p{1.46cm}rrrr|} 
\hline
\rowcolor{gray!10}
\textbf{WoW-Coh}  & \multicolumn{2}{r}{\textsl{Coherent Resp.}} & \textsl{Incoherent} & \textsl{All} \\
Train & \multicolumn{2}{r}{74,092} & 233,142       & 307,234 \\
Valid & \multicolumn{2}{r}{3,939} & 12,362        & 16,301  \\
Test & \multicolumn{2}{r}{3,865} & 12,098        & 15,963  \\
\hline
\hline
\rowcolor{gray!10}
\textbf{InferConv} & \textsl{Entail} & \textsl{Neutral} & \textsl{Contradict} & \textsl{All}\\
Train & 218,181 & 579,434 & 261,984 & 1,059,599\\
Valid & 28,072 & 12,242 & 9,780 & 50,094\\
Test & 3,119 & 1,087 & 1,360 & 5,566\\
\hline
\end{tabular}
\caption{Details of the datasets used to train our neural coherence scoring models.}
\label{Tab:sta2}
\vspace{-0.1in}
\end{table}

\begin{table}[t]
\setlength{\tabcolsep}{3pt}
\renewcommand{\arraystretch}{1.05}
\centering
\begin{scriptsize}
\begin{fontppl}
\begin{tabular}{lp{7cm}}
\toprule
\multicolumn{2}{l}{\textbf{Coherence}}\\
3 & TeacherBot provides coherent responses to the evaluator's input. \\
2 & TeacherBot provides largely coherent responses (with minor coherency issues) to the evaluator's input.  \\
1 & TeacherBot does not respond properly to the evaluator's input. \\
\midrule
\multicolumn{2}{l}{\textbf{Readability}}\\
3 & TeacherBot's responses are easy to read, containing no grammatical or semantic errors. \\
2 & TeacherBot's responses read smoothly but may contain 1-2 grammatical or semantic errors. \\
1 & TeacherBot's responses contain >2 grammatical or semantic errors, or are nonsensical.  \\
\midrule
\multicolumn{2}{l}{\textbf{Overall Quality}}\\
\multicolumn{2}{p{7.3cm}}{An initial score of 3 is given to a dialogue, then 1 point is deducted for each of the following issues, with a minimum score of 0.} \\
$\bullet$ & Uninformative, i.e. it provides <2 correct answers during QA. \\
$\bullet$ & Incoherent, i.e. the average coherence score is <2 points. \\
$\bullet$ & Low readability, i.e. the average readability score is <2.5 points. \\
$\bullet$ & Any other issues that could lead to an ineffective conversation e.g. words are repeated between turns. \\
\bottomrule
\end{tabular}
\end{fontppl}
\end{scriptsize}
\vspace{-0.05in}
\caption{
A scoring rubric provided to human evaluators.
}
\label{Tab:hum_score}
\vspace{-0.15in}
\end{table}

\begin{table*}[ht!]
\setlength{\tabcolsep}{4.5pt}
\renewcommand{\arraystretch}{1.05}
\begin{center}
\scriptsize
\begin{tabular}{| l | c | c c c | c c | c c c | c | c | c | c | c |} 
\hline
\multirow{2}{*}{\textbf{Dataset}}  & \multirow{2}{*}{\textbf{Model}}  & \multicolumn{5}{c|}{\textbf{Coverage Metrics}} & \multicolumn{3}{c|}{\textbf{Coherence Metrics}} & \multicolumn{4}{c|}{\textbf{Subjective Metrics}} & \multirow{2}{*}{Avg Len}  \\
\cline{3-14}
& & R-1  & R-2   & R-L   & QA$_{\textsc{Conf}}$   & QA$_{\textsc{F1}}$ & WoW-Coh & InferConv & DPR$_{\textsc{Relv}}$ &  QA$_{\textsc{Human}}$  & Coh & Read & Overall &  \\
\hline
\multirow{4}{*}{\textbf{Wikipedia}}     
 & \textsc{DoHA}    & 48.09 & 41.27 & 44.31 & 19.76 & 19.39 & 0.503 & 0.550 & 0.555 &  30.0 & 2.07 & 2.79 & 1.84 & 15.51 \\
 & \textsc{+Cov}  & \textbf{74.62} & \textbf{71.66} & \textbf{72.25} & \textbf{30.90} & \textbf{34.38} & 0.307 & 0.466 & 0.543  & \textbf{45.63} & 1.85 & 2.85 & 2.12 & 28.11 \\
 & \textsc{+Coh}  & 44.87 & 35.87 & 39.64 & 18.96 & 17.17 & \textbf{0.807} & \textbf{0.694} & \textbf{0.578} &  33.12 & \textbf{2.42} & 2.81 & 2.34  & 17.11\\
 & \textsc{Full}      & 62.78 & 58.83 & 60.61 & 25.94 & 27.01 & 0.617 & 0.630 & 0.556 & 38.74 & 2.26 & \textbf{2.88} & \textbf{2.37} & 20.69 \\
\hline
\multirow{4}{*}{\textbf{CNN-DM}}     
 & \textsc{DoHA}     & 38.89  & 30.09 & 32.91 & 15.90 & 15.80 & 0.521 &  0.567 & 0.538 & 28.12 & 2.28 & 2.67 & 1.87 & 16.29 \\
 & \textsc{+Cov}  & \textbf{81.52} & \textbf{78.52} & \textbf{73.73} & \textbf{30.98} & \textbf{38.46} &  0.253 & 0.381 & 0.525 & \textbf{58.59} & 2.14 & \textbf{2.79} & 2.37 & 36.08 \\
 &\textsc{+Coh}  & 30.45 & 18.61 & 24.0  & 13.31 & 11.31 & \textbf{0.845} & \textbf{0.692} & \textbf{0.561} & 40.15 & 2.40 & 2.48 & 1.93 & 16.23 \\
 & \textsc{Full}     & 65.77 & 60.45 & 57.83 & 25.5 & 30.36 & 0.604 & 0.692 & 0.559  & 52.5 & \textbf{2.57} & 2.71 & \textbf{2.53} & 29.76 \\
\hline
\multirow{4}{*}{\textbf{Papers}}     
& \textsc{DoHA}     & 36.27 & 28.20 & 30.60 & 10.29 & 5.34  & 0.565 & 0.452  & 0.557 & 30.5 & 1.58 & 2.55 & 1.4 & 15.61 \\
& \textsc{+Cov}  & \textbf{72.63} & \textbf{69.69} & \textbf{49.18} & \textbf{20.32} & \textbf{17.61}  & 0.271 & 0.141 & 0.529 & \textbf{57.5} & 1.73 & \textbf{2.77} & 1.96 & 33.55 \\
& \textsc{+Coh}  & 32.96 & 21.24 & 26.21 & 8.18  & 3.95  & \textbf{0.806}  & \textbf{0.547} & \textbf{0.576} & 25.47 & 1.82 & 2.26 & 1.25 & 16.3 \\
& \textsc{Full}      & 59.65 & 54.22 & 47.88 & 15.46 & 13.81 & 0.766 & 0.501 & 0.560 &  51.76 & \textbf{2.16} & 2.53 & \textbf{2.09} & 27.37 \\
\hline
\end{tabular}
\caption{
We compare Teacher bots based on naive DoHA\cite{prabhumoye-etal-2021-focused} model to variants fine-tuned using different reward functions, Avg len refers to average utterance length.}
\label{Tab:obj}
\end{center}
\vspace{-0.1in}
\end{table*}

\subsection{Comparison Models}
\label{sec:baselines}

Our baseline Teacher bot builds on the state-of-the art content-grounded dialogue generation model: \textsc{DoHA}~\cite{prabhumoye-etal-2021-focused}. 
It includes two improvements to the architectures of pre-trained encoder-decoder models~\cite{lewis-etal-2020-bart}: building context-driven representation of the supporting document, and enabling document-headed attention to acquire information from the document. 
\textsc{DoHA} has demonstrated strong performance in document-grounded generation. 
All \textsc{DoHA} models are pre-tuned on the \textsl{WoW} dataset.

Our \textsc{Full}  Teacher bot is created to converse in an informative and coherent manner. 
It extends \textsc{DoHA} by incorporating both coverage and coherence rewards in unsupervised self-play fine-tuning.
Additionally, we ablate \textsc{Full} model by removing each of the two rewards:
\textsc{+Cov} uses only the coverage reward for fine-tuning, i.e. setting $\beta=1$ in our reward function (\S\ref{Sec:rl}).
\textsc{+Coh} utilizes only the \textsl{WoW-coherence} reward, i.e. setting $\beta=0$.
Please refer to \hyperref[sec:appendix]{appendix} for
more implementation details and hyper-parameters.

\subsection{Evaluation Metrics}

We investigate a wide range of metrics to evaluate Teacher bot's performance.
\textsl{Objective metrics} measure the content coverage and coherence of Teacher bot's utterances.
\textsl{Subjective metrics}, devised with human-in-the-loop, provide a holistic evaluation of a conversation, focusing on its overall effectiveness and various aspects of linguistic quality. 

\begin{description}[style=unboxed,leftmargin=0cm]
\item[Objective Metrics.]
Teacher bot converses with Student bot over a passage for three turns.
That is, Teacher bot initiates the dialogue and provides two responses to Student bot. 
We objectively evaluate Teacher bot's utterances in terms of information coverage and coherence as follows.

\begin{itemize}[style=unboxed,leftmargin=0cm]
    \item 
\textsc{Rouge}~\cite{lin-2004-rouge} is one of the most widely used metrics for measuring information coverage.
We consider three variants in this study: \textsc{R-1}, \textsc{R-2} and \textsc{R-L}, which respectively measure the overlap of unigrams, bigrams and the longest common subsequence between  the given passage and Teacher bot's utterances.

\item  \textsl{QA}$_{\textsc{F1}}$ and \textsl{QA}$_{\textsc{Conf}}$ are two variants of 
SummaQA~\cite{scialom-etal-2019-answers}, a question answering-based evaluation metric.
If a conversation is rich in information, it could be used as a surrogate for the passage to answering important questions. 
To this end,
SummaQA generates Cloze-style questions from a passage by masking out entities, 
then employs a QA system to answer those questions based on a conversation. 
A higher QA performance suggests the conversation has better coverage. 
Particularly, \textsl{QA}$_{\textsc{F1}}$ reports the F1 score for question answering;
\textsl{QA}$_{\textsc{Conf}}$ measures the confidence of the QA system in predicting answers. 

\item  \textsl{WoW-Coherence} and \textsl{InferConv} are neural coherence scoring models (\S\ref{sec: coh}) repurposed for evaluation. 
These models quantitatively assess if Teacher bot has produced a coherent response given the conversation history, or not. 

\item \textsl{DPR$_{\textsc{Relv}}$} provides a new perspective on dialogue coherence evaluation~\cite{zhang2018reinforcing}.
It draws on the Dense Passage Retriever model (DPR; Karpukhin et al., 2020)\nocite{karpukhin-etal-2020-dense}
to predict if a Teacher bot's response is relevant to Student bot's input.
A higher relevance score means the input and response share the same topic, 
suggesting a coherent conversation.

\end{itemize}

\item[Subjective Metrics.]
We recruit 24 human evaluators to interact with Teacher bots. 
Each evaluator is asked to converse with bots over four passages. 
For each passage, the evaluator chats with four different Teacher bots for three turns,
where Teacher bot initiates the conversation and responds twice to the evaluator's input.
We randomly select 48 passages for evaluation, i.e., 16 passages from each of the three test sets.
To evaluate conversations produced from \textsl{Paper Abstracts}, we require evaluators, 8 in total, to be either PhD students or have obtained a PhD degree. 
For fair comparison, we shuffle and hide the order of Teacher bots presented to evaluators. 
Human evaluators were suggested to feed the same or similar inputs across Teacher bots on the same passage whenever possible. 
Throughout the conversation, the passage was not shown to the evaluators. 
After the conversation, human evaluators were asked to complete the following evaluation tasks: 

\begin{itemize}[style=unboxed,leftmargin=0cm]
\item \textsl{QA}$_{\textsc{Human}}$: 
Five sentences are randomly selected from each passage and one important entity is masked out in each sentence. The evaluators are presented with each corrupted sentence and asked if the sentence could be recovered by referencing the conversation with Teacher bot. We report the ratio of sentences that could be correctly recovered. 

\item \textsl{Linguistic Quality}: We ask human evaluators to rate each conversation along three dimensions: 
\emph{Coherence}: Does Teacher bot provide coherent responses to the evaluator's input?  
\emph{Readability}: Are Teacher bot's utterances easy to read, containing no grammatical or semantic errors?
\emph{Overall Quality}: How will the conversation score in terms of informativeness, coherence, readability and all aspects considered?
The scoring rubric provided to human evaluators is shown in Table~\ref{Tab:hum_score}.

\end{itemize}

\end{description}

\section{Objective Results}
\label{sec:obj_results}

\begin{description}[style=unboxed,leftmargin=0cm]
\item[Results on Test Sets.]
Table~\ref{Tab:obj} presents objective evaluation results obtained for various Teacher bots on three test sets: \textsl{CNN-DM}, \textsl{Paper Abstracts} and \textsl{Wikipedia}.
We observe that our \textsc{Full} Teacher bot is able to substantially outperform the baseline system \textsc{DoHA} on all datasets and across all objective metrics.  
It strikes a fine balance between delivering information-rich conversations and ensuring those conversations are coherent and attentive.  
Further, we find that optimizing a single reward, whether it be coverage or coherence, produces suboptimal results. 
For instance, \textsc{+Cov} tends to produce longer utterances than other variants. 
It improves information coverage, but yields low coherence scores,
leading to a performance even inferior to the baseline \textsc{DoHA}.
Our findings suggest that it is important for the reinforcer $L_{rl}^X$ to learn with both coverage or coherence rewards.

\begin{figure}[t]
\centering
\includegraphics[width=3.1in]{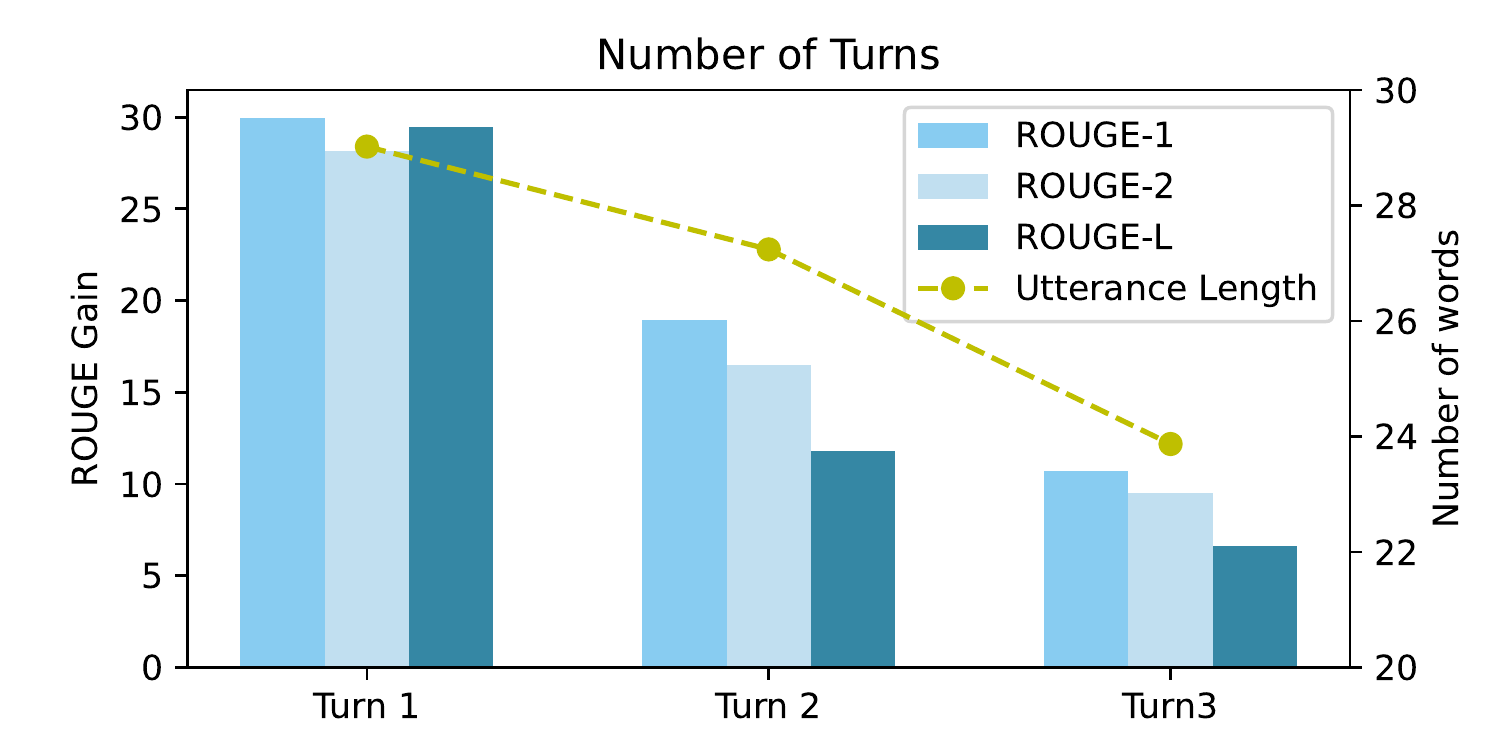}
\caption{ROUGE gain and utterance length tend to decrease as the number of turns increases}
\label{Fig: gain}
\end{figure}

\item[Trading off Coverage for Coherence.]
In Figure~\ref{Fig: finetune_scores}, we plot the learning curves of coverage and coherence scores when the reinforcer adopts a single reward (\textsc{+Cov}, \textsc{+Coh}) or both (\textsc{Full}). 
We use $R_{cov}$ and $R_{coh}$ to approximate coverage and coherence scores.
These plots are generated using 50 validation instances from the \textsl{Paper Abstracts} dataset.
We observe that with only the coverage reward (\textsc{+Cov}), Teacher bot tends to aggressively copy content from the passage, while disregarding the conversation history. 
This inevitably leads to incoherent conversations.
Conversely, \textsc{+Coh} can improve on coherence, but falls short on delivering informative conversations. 
Finally, our \textsc{Full} Teacher bot trades off coverage for substantially higher coherence, thus achieving a significant improvement over the baseline \textsc{DoHA} model.

\begin{figure*}
\centering
\includegraphics[width=5.5in]{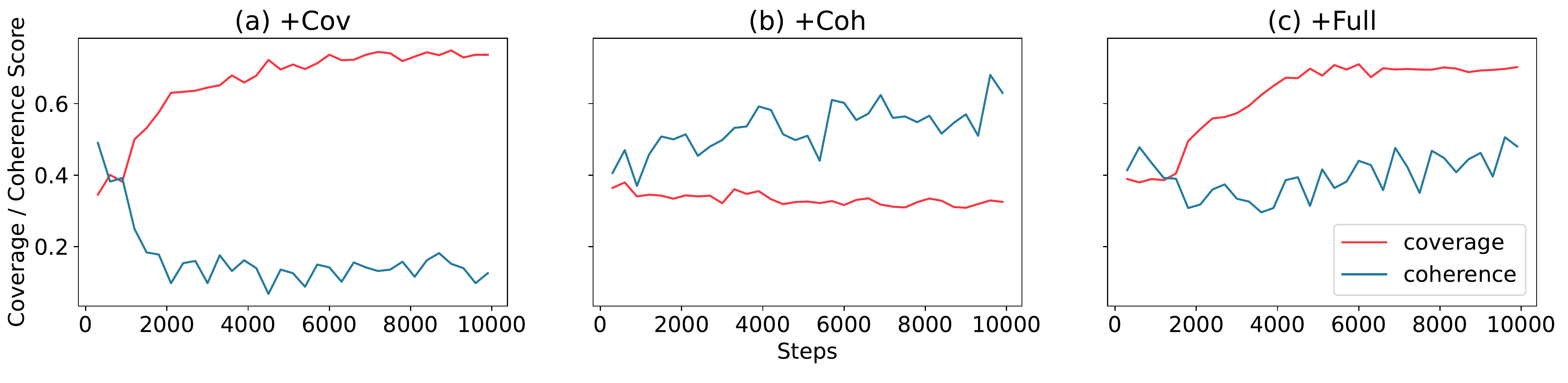}
\caption{Coverage and coherence scores when applying different reward scores}
\label{Fig: finetune_scores}
\end{figure*}

\item[Trading off Coverage for Coherence.]
We are curious to know the amount of information brought by each utterance produced by Teacher bot.  
To this end, we define information gain $\mathcal{IG}(\cdot)$ as the improvement of ROUGE scores brought by an utterance $U$:
\begin{align*}
\mathcal{IG}(U) = 
\textsc{Rouge}(P, H+U) - \textsc{Rouge}(P, H),
\end{align*}

\noindent where $P$ is the supporting passage, and $H$ represents the conversation history.
We consider three \textsc{Rouge} variants, \textsc{R-1}, \textsc{R-2} and \textsc{R-L}, respectively.
Figure~\ref{Fig: gain} illustrates the gain of information for each of the three turns.
The average \textsc{Rouge} gain is reported for each turn, using conversations produced for the \textsl{Paper Abstracts} dataset. 
We observe that there is a general tendency across turns that information gain is decreasing. 
This is in part because that at the beginning of a conversation, Teacher bot has no constraints regarding content selection, it could rephrase any content selected from the supporting passage to initiate a dialogue. 
In subsequent turns, Teacher bot has to exercise caution in response generation considering both the conversation history and overall coherence of the conversation. 
Consequently, we find that the average length of the utterances also decreases in subsequent turns.

\begin{table}[t]
\setlength{\tabcolsep}{4.5pt}
\renewcommand{\arraystretch}{1.05}
\begin{center}
\scriptsize
\begin{tabular}{| l | c | c c | c c c |} 
\hline
\multirow{2}{*}{\textbf{Dataset}}  & \multirow{2}{*}{\textbf{Model}}  & \multicolumn{2}{c|}{\textbf{Coverage}} & \multicolumn{3}{c|}{\textbf{Coherence}}  \\
\cline{3-7}
& & \textsl{QA}$_{\textsc{Conf}}$ & \textsl{QA}$_{\textsc{F1}}$ & WoW-Coh & IC & DPR  \\
\hline
\multirow{2}{*}{\textbf{Wikipedia}}     
& \textsl{WoW}  & 18.96 & 17.17 & \textbf{0.807} & \textbf{0.694} & \textbf{0.578} \\
& \textsl{InferCov}   & 20.41 & 19.61 & 0.647  & 0.676  & 0.561 \\
\hline
\multirow{2}{*}{\textbf{CNN-DM}}     
& \textsl{WoW}  & 13.31 & 11.31  & \textbf{0.845} & 0.692 & \textbf{0.561} \\
& \textsl{InferCov}   & 14.95 & 14.07  & 0.719 & \textbf{0.728} & 0.550 \\
\hline
\multirow{2}{*}{\textbf{Papers}}     
& \textsl{WoW}  & 8.18  &  3.95 & \textbf{0.806} & 0.547 & \textbf{0.576} \\
& \textsl{InferCov}   & 9.47  &  4.31 & 0.737 & \textbf{0.554} & 0.558 \\
\hline
\end{tabular}
\caption{Teacher bot's performance when using different coherence score models during fine-tuning}
\label{Tab:obj-coh}
\end{center}
\vspace{-0.1in}
\end{table}

\item[A Comparison of Coherence Scoring Models.]
We compare two Teacher bots fine-tuned only with coherence reward from different coherence scoring models (i.e. \textsl{WoW-coherence} and \textsl{InferConv}). We demonstrate their objective results in Table~\ref{Tab:obj-coh}. According to the results,  Teacher bot fine-tuned with \textsl{InferConv} achieves slightly better coverage metrics. However, in terms of coherence metrics, Teacher bot fine-tuned with \textsl{WoW-coherence} model generally achieves better performance. Based on this observation, \textsl{WoW-coherence} scoring model better measures coherence in conversations.

\end{description}

\section{Subjective Results}

We demonstrate the subjective evaluation results in Table~\ref{Tab:obj} and have the following observations: 
\begin{enumerate}[style=unboxed,leftmargin=0cm]
    \item 
 For question answering, \textsc{+Cov} achieves the best performance on all three datasets. This again proves that the coverage reward helps make the conversation more informative;

\item For coherence scores, \textsc{+Coh} achieves the best performance on \textsl{Wikipedia}. However, on the other two datasets it was outperformed by \textsc{Full}. 

\item For readability scores, on \textsl{CNN-DM} and \textsl{Paper Abstracts}, \textsc{+Coh} achieves the lowest performance while \textsc{+Cov} achieves the highest.  

\item For overall scores, \textsc{Full} demonstrates the best performance. This suggests \textsc{Full} delivers conversations that are more balanced in coverage, coherence and readability. 

\item The \textsl{Paper Abstracts} corpus is the most challenging among all the test corpora, as our Teacher bots generally show worse performance in coherence, readability and overall scores. We found passages in \textsl{Paper Abstracts} contain volumes of professional vocabularies thus are more complicated for people to understand. In addition, it's also more difficult for Student bot to respond appropriately during self-play fine-tuning (See examples in \hyperref[sec:appendix]{appendix}). As a result, transferring Teacher bots to this domain is more challenging.

\end{enumerate}

\begin{table}[h]
\setlength{\tabcolsep}{3pt}
\renewcommand{\arraystretch}{1.05}
\centering
\begin{scriptsize}
\begin{fontppl}
\begin{tabular}{l p{6.5cm}}
\toprule[2pt]
\multicolumn{2}{c}{\textbf{Example A} (Passage from \textsl{Wikipedia})}\\
\midrule
\rowcolor{Gray}{\textbf{Passage}} & The Concorde is a British–French turbojet-powered supersonic passenger airliner ... It had a maximum speed over twice the speed of sound, at Mach 2.04, ... First flown in 1969, Concorde entered service in 1976 and operated for 27 years. ... \\
{\textbf{User}} & What is the plane famous for? \\

\rowcolor{Gray}{\textbf{Bot Cov}} &  First flown in 1969, Concorde entered service in 1976 and operated for 27 years.  \\

{\textbf{Bot Coh}}&   Concorde is famous for \underline{the speed of sound} at Mach 2.04. \\

\rowcolor{Gray}{\textbf{Bot Full}}&  Concorde is famous for having a maximum speed over twice the speed of sound, at Mach 2.04. \\
\midrule

\multicolumn{2}{c}{\textbf{Example B} (Passage from \textsl{CNN News})}\\
\midrule
\rowcolor{Gray}{\textbf{Passage}} & The US Food and Drug Administration, ... Molnupiravir is for the treatment of mild-to-moderate coronavirus disease in adults with positive results of direct SARS-CoV-2 viral testing ... This is the second Covid-19 antiviral pill authorized for ill people to take at home, before they get sick enough to be hospitalized ... \\

{\textbf{User}} & Tell me more about the drug. \\

\rowcolor{Gray}{\textbf{Bot Cov}}  & This is the second Covid-19 antiviral pill authorized for ill people to take at home, before they get sick enough to be hospitalized.  \\

{\textbf{Bot Coh}}& The drug is for people with mild to moderate coronavirus disease. \\

\rowcolor{Gray}{\textbf{Bot Full}}& Molnupiravir is for the treatment of mild-to moderate coronavirus disease in adults with positive results of direct SARS-CoV-2 viral testing. \\
\midrule

\multicolumn{2}{c}{\textbf{Example C} (Passage from \textsl{Paper Abstracts})}\\
\midrule
\rowcolor{Gray}{\textbf{Passage}} &  ... In this paper we describe a novel framework to guide a supervised, extractive summarization system with question-answering rewards ... Our results compare favorably with those reported by strong summarization baselines as evaluated by automatic metrics and human assessors ... \\
{\textbf{User}} & What did you propose? \\

\rowcolor{Gray}{\textbf{Bot Cov}} &  Our results compare favorably with those reported by strong summarization baselines as evaluated by automatic metrics and human assessors.  \\

{\textbf{Bot Coh}}&  A proposal to use the extractive summarizer \underline{to answer questions.} \\

\rowcolor{Gray}{\textbf{Bot Full}}& We propose a novel framework to guide a supervised, extractive summarization system with question-answering rewards. \\
\bottomrule[2pt]
\end{tabular}
\end{fontppl}
\end{scriptsize}
\vspace{-0.05in}
\caption{A few example responses from different Teacher bots. 
In example A,\textsc{+Coh} and \textsc{Full} abstractively generates \emph{Concord is famous for ...} to make the response more coherent to user's question. However, the \underline{underlined part} in \textsc{+Coh}'s statement misrepresent the input passage and is inaccurate.
In example B, all responses seem coherent because of the open-ended question.
}
\label{Tab:case}
\vspace{-0.15in}
\end{table}

\begin{description}[style=unboxed,leftmargin=0cm]
\item[A Case Study.]

\noindent We show a few Teacher bots' responses to users in Table~\ref{Tab:case}. After analyzing the cases, we have the following observations: 
\begin{enumerate}[style=unboxed,leftmargin=0cm]
    \item 
The coverage reward encourages Teacher bots to directly copy content from the input passage, while the coherence reward encourages  abstractively generating new content: As shown in Example A, \textsc{+Cov} directly extracts a part of the original passage as response, regardless of the user's question, while \textsc{+Coh} and \textsc{Full} abstractively rewrite the response to make it more coherent. 

\item
Putting too much weight on coherence reward could make Teacher bot become so abstractive that it misrepresents the original passage and lead to incoherence and semantic/grammar errors. (See \textsc{+Coh}'s response in example A and C) This explains \textsc{+Coh}'s low coherence and readability scores on \textsl{CNN-DM} and \textsl{Paper Abstracts}. This observation suggests the necessity to carefully choose the weight for coherence rewards, and to coupling coherence reward with coverage reward, which could make the chat bot less abstractive. 

\item
Generally, user utterances could be classified into two categories: \textbf{Information-seeking queries}  which request certain information (e.g. the user's utterance in example A); \textbf{Open statements} which do not have specific requests (e.g. the user's utterance in example B). We found evaluators tend to give high coherence scores to response to open statements, as they could be addressed by a wider range of responses.

\end{enumerate}

\end{description}

\section{Related Work}

\begin{description}[style=unboxed,leftmargin=0cm]
\item[Content-Grounded Dialogue Generation]
Content grounded dialogue generation is the task of using the information provided in external content (e.g. a passage, etc.) to guide dialogue generation. 

Compared to previous research, our task has the following novelties.
1) Compared to content-grounded information-retrieval-oriented dialogue such as doc2dial~\cite{, feng-etal-2020-doc2dial} and ABCD~\cite{chen-etal-2021-action} where the chat bot responds to user query in a passive way, we expect our chat bots to convey knowledge proactively. 
2) Compared to chit chat-oriented dialogue such as~\cite{zhou-etal-2018-dataset, dinan2019wizard, blenderbot1Komeili, blenderbot2Xu}, our task is more focused on extensive conversation in a particular topic, and aims at helping the end user acquire knowledge or information from a given passage. 
3) Contrasted to chat bots that are applied in a single domain~\cite{zhou-etal-2018-dataset, moghe-etal-2018-towards, xu2019review}, our chat bot could be transferred to other domains through self-talk based fine-tuning. 

Another line of research works focus on content grounded text generation models~\cite{prabhumoye-etal-2021-focused, Zhao2020Low-Resource}. Compared with ordinary text generation models (e.g. BART~\cite{lewis-etal-2020-bart}), these models are specifically designed to model external content as an additional input, and achieve better performance on content grounded dialogue generation tasks including \textsl{CMU DoG}~\cite{Zhao2020Low-Resource} and \textsl{Wizard of  Wikipedia}~\cite{dinan2019wizard}.

\item[RL in Text Generation]
Reinforcement learning has been applied in various natural language generation tasks, including image caption~\cite{Rennie2017cvpr}, automatic summarization~\cite{paulus2018a}, machine translation~\cite{kang-etal-2020-dynamic} and poem generation~\cite{yang-etal-2019-generating}. Specifically, when applying reinforcement learning in dialogue generation~\cite{li-etal-2016-deep, zhao-etal-2019-rethinking, shi-etal-2019-build, yamazaki-aizawa-2021-phrase, liu-etal-2020-impress}, self-play is often used to enable scoring multi-turn dialogues.
Compared to previous dialogue generation research using RL and self-play, our two-phase framework enables transferring the teacher bot to other domains by optimizing a mixed reward of coverage and coherence.

\item[Educational Dialogue Systems and Conversational QA]

There have been research works applying dialogue systems for educational purposes.
Some chat bots are for language practice.~\cite{Ruan2021EnglishBotAA, Stewart2007chat, Huang2017chat}
Others are specially designed for education in a single domain or task, e.g. moral education~\cite{peng2019task}, educational debate~\cite{yuan2008human}. Compared with previous educational dialogue systems, our system is for information acquisition without domain restriction. 
Our task is also related to conversational question answering (CQA), e.g.~\cite{kocisky-etal-2018-narrativeqa, rajpurkar2016squad, joshi2017triviaqa, zellers-etal-2018-swag}. However, most existing CQA systems passively response to user queries in single turn conversations, while our system actively engage with users in multi-turn conversations. 

\end{description}

\section{Conclusion}

We propose an information-acquisition-oriented dialogue system that transfers information and knowledge in passages to users through conversations.  
An unsupervised self-talk approach is introduced leveraging novel rewards to enable Teacher bots to deliver informative and attentive conversations. 
Experiments with automatic and human evaluations demonstrate the effectiveness of our approach. Some interesting future directions include extending the conversations to be based on a set of documents and specializing our dialogue systems for in specific domain, e.g. patient education.

\section{Ethical Considerations}

Our models are pre-tuned on \textsl{Wizard of Wikipedia} dataset and fine-tuned on three corpora: \textsl{Wikipedia}, \textsl{CNN-DailyMail} and \textsl{Paper-Abstracts} (Abstracts of papers from ACL, EMNLP, NAACL, EACL, Findings and ICLR submissions from 2017 to 2021). All the datasets used in this paper are publicly available. Moreover, we did not use full-length Wikipedia or CNN-Daily Mail news articles in our experiments, but tailored versions of 100-150 words. This is because a full length Wikipedia/CNN-Daily Mail article may contain too much content to be covered in a short conversation.

As described in~\cite{maynez-etal-2020-faithfulness,kryscinski-etal-2020-evaluating,lebanoff-etal-2020-understanding,zhou-etal-2021-detecting}, current state of the art neural conditional text models can output hallucinated content unfaithfully to the input text, which impedes the safe deployment of the models. We note that our Teacher bots may also generate utterances that are not supported by the input passage. 

\newpage

\bibliography{0_acl_latex}

\bibliographystyle{acl_natbib}

\newpage
\section{Appendix}
\label{sec:appendix}

\subsection{Implementation Details}

\subsubsection{Pre-tuning}

\noindent\textbf{Key hyper-parameters for both models. }

\begin{itemize}[topsep=0pt, partopsep=0pt, itemsep=0pt,parsep=0pt]

  \item source max len: 1024
  \item target max len: 128
  \item batch size: 8
  \item train epoch: 3
  \item learning rate: 2e-5
  \item Both Teacher and Student bots adopt the initialized weights of \href{https://huggingface.co/facebook/bart-base}{bart-base} from huggingface.
  \item Both DoHA models and BART models are based on the implementation presented in the DoHA paper~\cite{prabhumoye-etal-2021-focused}.
\end{itemize}

Note we initialize our models with \href{https://huggingface.co/facebook/bart-base}{bart-base} instead of bart-large as our self-play fine-tuning is very computational intensive and time consuming. With our current setting, the self-play fine-tuning takes about 2.5 days on one single NVIDIA Tesla V100 GPU. 

\subsubsection{Self-Play Reinforced Fine-Tuning}

\noindent\textbf{Mixed Loss. }Optimizing our mixed loss is achieved by intermittently minimizing the MLE loss for a $a$ batches and then minimizing the RL loss for $b$ batches. In other word, the parameter $\gamma$ in Section~\ref{Sec: usp} is determined by $a$ and $b$. Specifically, in each fine-tuning step, the parameters of Teacher bot are updated for 4 times, once over an RL batch, 3 times over MLE batches. 

\noindent\textbf{Key Hyper-Parameters. }We apply the same set of basic hyper-parameters for all Teacher bots during all fine-tuning process:
\begin{itemize}[topsep=0pt, partopsep=0pt, itemsep=0pt,parsep=0pt]
  \item MLE batch size: 8
  \item RL batch size: 5
  \item maximum coverage score: 0.5
  \item train epoch: 3
  \item learning rate: 1e-6
  \item Fine-tuning steps: 10,000
  \item $\beta$ for three reinforced Teacher bots: \textsc{Full}-$\beta=0.7$, \textsc{+Cov}-$\beta=1.0$, \textsc{+Coh}-$\beta=0.0$
\end{itemize}

\subsection{Coherence Scoring Models}

Both \textsl{WoW-coherence} and \textsl{InferConv} are trained based on \hyperlink{https://huggingface.co/bert-base-uncased}{bert-base-cased}. 

\noindent\textbf{Key hyper-parameters for both models. }

\begin{itemize}[topsep=0pt, partopsep=0pt, itemsep=0pt,parsep=0pt]
  \item max length: 256
  \item batch size: 32
  \item learning rate: 2e-5
  \item epochs: 3
\end{itemize}

For \emph{InferConv} classification model, the constant scores are $s_e=1.0$, $s_n=0.2$ and $s_c=0.0$

The \emph{WoW-coherence} and \emph{InferConv} classification model achieves 82.1\% and  88.4\% accuracy on respective test sets. 

All other hyper-parameter settings for Teacher bot, Student bot and coherence scoring models are based on the system's default setting. All our experiments were run on servers with Nvidia A100 and V100 GPUs.

\subsection{Dataset Details}

The passages in train/validation/test set of our \textsl{Wikipedia} corpus are randomly sampled from passages in train/validation/test set of \textsl{WoW} respectively. Similarly, passages in train/validation/test set of our \textsl{CNN-DM} corpus are randomly sampled from the train/validation/test set of the original \emph{CNN-DM} dataset. For \textsl{Paper Abstracts}, we randomly distribute all collected paper abstracts into train/validation/test sets.

\subsection{Rewards function Details}

During our exploration, we have explored multiple variations of coverage reward functions: 

\begin{enumerate}
  \item Reward score is gained at the end of each turn of conversation, and it is calculated the ROUGE score improvement of the teacher bot's utterance (As we applied in the paper);
  \item Reward score is gained at the end of each turn of conversation, and it is calculated the ROUGE score improvement of both the teacher bot and student bot's utterances;
  \item Reward score is gained at the end of the entire conversation, and is calculated as the ROUGE score improvement of all teacher bot's utterances. 
\end{enumerate}

According to our experience, the performance of 1 is similar than 2. Specifically, we notice in most cases, our student bot contribute marginally in terms of ROUGE score improvement. In addition, in practice, we observe 1 shows better performance than 3.

\subsection{Human Evaluation Details}

 All human evaluation passages are collected randomly, the length of the \textsl{Wikipedia} and \textsl{CNN} passages are tailored to 100-150 words to conform to the length of passages in the fine-tuning stage.  During conversations, we suggest our evaluators to use utterances relevant to the topic, so that their utterances could be appropriately addressed by Teacher bots referencing content in the passage. All human evaluators we recruit have at least a bachelor degree, each evaluator is rewarded with a \$20 gift card for participation.

The evaluation results are collected using Google Colab, an example evaluation page is available  \href{https://colab.research.google.com/drive/1bXhtCzARyHFmcd4bXZak04gm7HZ42Wdt?usp=sharing}{through this link} (Personal information are anonymized during reviewing stage).

Note that in our human evaluation, Teacher bot chat about a passage with two evaluators. We use Pearson correlation coefficient to measure Inter-annotator agreement, the average Pearson correlation coefficient among each pair of evaluators is 0.16, implying weakly positive correlation. We note different evaluators use different utterances to chat with the same Teacher bot, thus it is reasonable that the conversation with the same Teacher bot over the same passage have large variance. 

\subsection{Examples}

We provide more examples of conversations in self-play in Table~\ref{Tab:case_stb} and human evaluations in Table~\ref{Tab:case_usr}.

\begin{table*}[t]
\setlength{\tabcolsep}{3pt}
\renewcommand{\arraystretch}{1.05}
\begin{footnotesize}
\begin{fontppl}
\centering

\begin{tabular}{l p{13.9cm}}

\toprule[2pt]
\multicolumn{2}{c}{\textbf{Example D} (Passage from \textsl{Wikipedia})}\\
\midrule
\rowcolor{Gray}{\textbf{Passage}} & For 4,000 years China has been a nation of farmers. By the time the People's Republic of China was established in 1949, virtually all arable land was under cultivation; irrigation and drainage systems constructed centuries earlier and intensive farming practices already produced relatively high yields. But little prime virgin land was available to support population growth and economic development. However, after a decline in production as a result of the Great Leap Forward (1958–60), agricultural reforms implemented in the 1980s increased yields and promised even greater future production from existing cultivated land. \\
\hline
{\textbf{Dialogue}} & \textbf{[Teacher bot]:} China has been a nation of farmers for 4,000 years! \\
                    & \textbf{[Student bot]:} Wow, that's a long time. What do they grow? \\
                    & \textbf{[Teacher bot]:} They grow a variety of crops, including chicken, fish, fruits, and vegetables.  \\
                    & \textbf{[Student bot]:} That sounds like a lot of food. What else can you tell me about China?   \\
                    & \textbf{[Teacher bot]:} Well, by the time the People's Republic of China was established in 1949, virtually all arable land was under cultivation.   \\
\midrule
\multicolumn{2}{c}{\textbf{Example E} (Passage from \textsl{CNN-DailyMail})}\\
\midrule
\rowcolor{Gray}{\textbf{Passage}} & Everywhere you look there are two-wheeled drivers waiting ominously for passengers to hop on board. They're the backbone of public transport in Uganda and the fastest way to get around the capital city, Kampala. The boda boda motorbike taxi is a staple used by all sectors of society. Traditionally, the usual way of catching a ride was to venture into any street corner packed with tens of boda bodas or simply waiting for one to pass by. But now, a new local startup are bringing this classic mode of transport into the technological age by providing the ability to hail a boda boda at the click of a smartphone. "Bodas are the main thing getting people from A to B," says Alastair Sussock, co-founder of SafeBoda ... \\

{\textbf{Dialogue}} & \textbf{[Teacher bot]:} Ugandan start-up SafeBoda are bringing this classic mode of transport into the technological age by providing the ability to hail a boda boda at the click of a smartphone.  \\
                    & \textbf{[Student bot]:} I've never heard of them. What are they? \\
                    & \textbf{[Teacher bot]:} They are the backbone of public transport in Uganda and the fastest way to get around the capital city, Kampala.   \\
                    & \textbf{[Student bot]:} That's really cool. How long have they been around?  \\
                    & \textbf{[Teacher bot]:} Well they have been around for a long time. Traditionally, the usual way of catching a ride was to venture into any street corner packed with tens of boda bodas or simply waiting for one to pass by.   \\
\midrule
\multicolumn{2}{c}{\textbf{Example F} (Passage from \textsl{Paper Abstracts})}\\
\midrule
\rowcolor{Gray}{\textbf{Passage}} & Existing open-domain question answering (QA) models are not suitable for real-time usage because they need to process several long documents on-demand for every input query, which is computationally prohibitive. In this paper, we introduce query-agnostic indexable representations of document phrases that can drastically speed up open-domain QA. In particular, our dense-sparse phrase encoding effectively captures syntactic, semantic, and lexical information of the phrases and eliminates the pipeline filtering of context documents. Leveraging strategies for optimizing training and inference time, our model can be trained and deployed even in a single 4-GPU server. Moreover, by representing phrases as pointers to their start and end tokens, our model indexes phrases in the entire English Wikipedia (up to 60 billion phrases) using under 2TB. Our experiments on SQuAD-Open show that our model is on par with or more accurate than previous models with 6000x reduced computational cost, which translates into at least 68x faster end-to-end inference benchmark on CPUs ... \\

{\textbf{Dialogue}} & \textbf{[Teacher bot]:} We introduce query-agnostic indexable representations of document phrases that can drastically speed up open-domain QA. \\
                    & \textbf{[Student bot]:} What is QA? \\
                    & \textbf{[Teacher bot]:} It is a process where our dense-sparse phrase encoding effectively captures syntactic, semantic, and lexical information of the phrases and eliminates the pipeline filtering of context documents.    \\
                    & \textbf{[Student bot]:} What are the basic concepts of QA?   \\
                    & \textbf{[Teacher bot]:} Well the basic concepts are that we represent phrases as pointers to their start and end tokens, our model indexes phrases in the entire English Wikipedia (up to 60 billion phrases) using under 2TB. \\
\bottomrule[2pt]

\end{tabular}

\end{fontppl}
\end{footnotesize}
\vspace{-0.05in}
\caption{Conversation examples of Teacher bot fine-tuned with both rewards chat with Student bot during self-play. Example F demonstrates an example where Student bot's question seems too broad to be responded using content in the passage.}
\label{Tab:case_stb}
\vspace{-0.15in}
\end{table*}

\begin{table*}[t]
\setlength{\tabcolsep}{3pt}
\renewcommand{\arraystretch}{1.05}
\begin{footnotesize}
\begin{fontppl}
\centering

\begin{tabular}{l p{13.9cm}}

\toprule[2pt]
\multicolumn{2}{c}{\textbf{Example G} (Passage from \textsl{Wikipedia})}\\
\midrule
\rowcolor{Gray}{\textbf{Passage}} & Interstellar is a 2014 epic science fiction film co-written, directed and produced by Christopher Nolan. It stars Matthew McConaughey, Anne Hathaway, Jessica Chastain, Bill Irwin, Ellen Burstyn, and Michael Caine. Set in a dystopian future where humanity is struggling to survive, the film follows a group of astronauts who travel through a wormhole near Saturn in search of a new home for mankind. Paramount Pictures, Warner Bros. Pictures, and Legendary Pictures co-financed the film. Cinematographer Hoyte van Hoytema shot it on 35 mm in the Panavision anamorphic format and IMAX 70 mm. Interstellar uses extensive practical and miniature effects and the company Double Negative created additional digital effects. Interstellar premiered on October 26, 2014, in Los Angeles, California. In the United States, it was first released on film stock, expanding to venues using digital projectors. The film had a worldwide gross of over \$677 million (and \$701 million with subsequent re-releases), making it the tenth-highest-grossing film of 2014.
 \\
\hline
{\textbf{Dialogue}} & \textbf{[Teacher bot]:} Interstellar is a 2014 epic science fiction film directed and produced by Christopher Nolan. \\
                    & \textbf{[Human user]:} Who stars in the film? \\
                    & \textbf{[Teacher bot]:} The film stars Matthew McConaughey, Anne Hathaway, Jessica Chastain, Bill Irwin, Ellen Burstyn, and Michael Caine.  \\
                    & \textbf{[Human user]:} What story does it tell?   \\
                    & \textbf{[Teacher bot]:} The film follows a group of astronauts who travel through a wormhole near Saturn.   \\
\midrule
\multicolumn{2}{c}{\textbf{Example H} (Passage from \textsl{CNN News})}\\
\midrule
\rowcolor{Gray}{\textbf{Passage}} & The US Food and Drug Administration on Monday granted full approval to the Pfizer/BioNTech Covid-19 vaccine for people age 16 and older. This is the first coronavirus vaccine approved by the FDA, and is expected to open the door to more vaccine mandates. The vaccine will be marketed as Comirnaty, the FDA said in its announcement on Monday. The Pfizer/BioNTech vaccine has been authorized for emergency use in the United States since mid-December for people age 16 and older, and in May, the authorization was extended to those 12 and older. FDA says it's working as fast as possible to fully approve vaccines, as urgency rises amid Covid surge FDA says it's working as fast as possible to fully approve vaccines, as urgency rises amid Covid surge ... \\

{\textbf{Dialogue}} & \textbf{[Teacher bot]:} The Pfizer/BioNTech vaccine has been authorized for emergency use in the United States since mid-December for people age 16 and older.  \\
                    & \textbf{[Human user]:} Give me more details. \\
                    & \textbf{[Teacher bot]:} This is the first coronavirus vaccine approved by the FDA, and is expected to open the door to more vaccine mandates.   \\
                    & \textbf{[Human user]:} So, what will happen next?  \\
                    & \textbf{[Teacher bot]:} The Pfizer/BioNTech Covid-19 vaccine for people age 16 and older will be marketed as Comirnaty, the FDA said in its announcement on Monday.  \\
\midrule
\multicolumn{2}{c}{\textbf{Example I} (Passage from \textsl{Paper Abstracts})}\\
\midrule
\rowcolor{Gray}{\textbf{Passage}} & We introduce a new language representation model called BERT, which stands for Bidirectional Encoder Representations from Transformers. Unlike recent language representation models, BERT is designed to pre-train deep bidirectional representations from unlabeled text by jointly conditioning on both left and right context in all layers. As a result, the pre-trained BERT model can be fine-tuned with just one additional output layer to create state-of-the-art models for a wide range of tasks, such as question answering and language inference, without substantial task-specific architecture modifications.
BERT is conceptually simple and empirically powerful. It obtains new state-of-the-art results on eleven natural language processing tasks, including pushing the GLUE score to 80.5\% (7.7\% point absolute improvement), MultiNLI accuracy to 86.7\% (4.6\% absolute improvement), SQuAD v1.1 question answering Test F1 to 93.2 (1.5 point absolute improvement) and SQuAD v2.0 Test F1 to 83.1 (5.1 point absolute improvement).\\

{\textbf{Dialogue}} & \textbf{[Teacher bot]:} We introduce BERT, which stands for Bidirectional Encoder Representations from Transformers. \\
                    & \textbf{[Human user]:} What did you propose? \\
                    & \textbf{[Teacher bot]:} A new language representation model called BERT which is conceptually simple and empirically powerful.    \\
                    & \textbf{[Human user]:} How is the result?   \\
                    & \textbf{[Teacher bot]:} The pre-trained BERT model can be fine-tuned with just one additional additional output layer to create state-of-the-art models for a wide range of tasks, such as question answering and language inference, without substantial task-specific architecture modifications. \\
\bottomrule[2pt]

\end{tabular}

\end{fontppl}
\end{footnotesize}
\vspace{-0.05in}
\caption{Conversation examples of Teacher bot fine-tuned with both rewards chat with human evaluators.}
\label{Tab:case_usr}
\vspace{-0.15in}
\end{table*}

\end{document}